\documentclass[10pt,twocolumn]{article}

\usepackage{3dv}
\usepackage{times}
\usepackage{epsfig}
\usepackage{graphicx}
\usepackage{amsmath}
\usepackage{amssymb}
\usepackage{bm}
\RequirePackage{hyperref}
\RequirePackage{subcaption}

\threedvfinalcopy 

\ifthreedvfinal\fi
\begin{document}

\title{Fast, Accurate and Object Boundary-Aware Surface Normal Estimation from Depth Maps}

\author{Saed Moradi\\
Computer Vision and Systems Laboratory (CVSL)\\ Laval University
Quebec, QC G1V0A6, Canada
\and
Alireza Memarmoghadam\\
Department of Electrical Engineering\\University of Isfahan, Isfahan, Iran
\and
Denis Laurendeau\\
Computer Vision and Systems Laboratory (CVSL)\\ Laval University
Quebec, QC G1V0A6, Canada
}

\maketitle
\thispagestyle{empty}

\begin{abstract}
   This paper proposes a fast and accurate surface normal estimation method which can be directly used on depth maps (organized point clouds). The surface normal estimation process is formulated as a closed-form expression. In order to reduce the effect of measurement noise, the averaging operation is utilized in multi-direction manner. The multi-direction normal estimation process is reformulated in the next step to be  implemented efficiently. Finally, a simple yet effective method is proposed to remove erroneous normal estimation at depth discontinuities.  The proposed method is compared to well-known surface normal estimation algorithms. The results show that the proposed algorithm not only outperforms the baseline algorithms in term of accuracy, but also is fast enough to be used in real-time applications. 
\end{abstract}

\section{Introduction}

Surface normal vectors estimation \cite{lenssen2020deep,do2020surface,seo20203d} is the common process in different 3D vision and 3D processing task such as 3D surface reconstruction \cite{lu2020deep}, scene segmentation \cite{poux2020unsupervised}, object recognition \cite{zhao2020hoppf}, and others.  A complete 3D processing pipeline may fail due to lack of effective surface normal estimation process. Thus,  fast and accurate normal estimation is of great importance in practical application. 
There are several approaches dedicated to normal estimation in the literature depending on the type of input 3D data. For an unorganized point clouds (an unordered set of 3D points),  at the first step, a graph should be constructed to identify the neighboring points for each query points. Then, the vector normal to the query point is estimated. The most common approach for surface normal estimation is called plane-PCA (\cite{berkmann1994computation}). In this method, a plane is fitted to the all neighboring points (including the query point). Then, the vector normal to the fitted plane is determined by eigen decomposition of the scattering matrix. The eigen vector correspond to the smallest eigen value is considered as the surface normal vector. The plane-PCA method is a robust and accurate method. However, it is not suitable for large-scale point clouds due to its high computational complexity. Beside the plane-PCA, too many research have been done to improve the accuracy of the surface normal estimation process.

 In \cite{li2010robust}, in order to improve the normal estimation accuracy for points belonging to high curvature surfaces, an optimum tangent plane is fitted using robust statistics. Randomized Hough Transform (RHT) along with statistical exploration bounds is used to preserve sharp features in  \cite{boulch2012fast}. To reduce the computational complexity, the authors used a fixed-size accumulator. A GPU-based implementation is proposed in \cite{liu2012normal} to speed-up a computationally intensive tensor voting algorithm. 
 In order to exclude the outliers, the Deterministic MM-estimator (DetMM) is used in \cite{khaloo2017robust}. 
 In addition to classical data processing techniques, deep learning-based methods have recently attracted the attention of the research community for surface normal vector estimation \cite{lenssen2020deep,zhou2021improvement}. However, these methods usually require richly labeled datasets. 

Comparing to unorganized point clouds, surface normal estimation directly from a depth map (organized point cloud) has received less attention in the literature. However, estimating surface normal  from depth maps has the following advantages: 
\begin{itemize}
\item No extra processing step is needed to determine the points belonging to the neighborhood of a query point.
\item The complete normal estimation process can be implemented through 2D image processing operators which are much faster than 3D ones.
\end{itemize}

 Some research in the literature  focuses on the estimation of  normal vectors directly from input depth images. In \cite{tang2012histogram} a closed-form expression is proposed for estimating normal vectors at each point. However, improper tangent vector selection led to inaccurate normals map.  Holzer et al. \cite{holzer2012adaptive} proposed a surface normal estimation method based on adaptive neighboring size selection and integral images. However,  the accuracy of their method depends on hyperparameters values which are chosen empirically.  Also, the performance of their method is degraded facing small objects with high surface curvature. One of the most accurate and fast surface normal estimation methods is presented in  \cite{nakagawa2015estimating}. While the surface tangent vectors are constructed perfectly, the approximation error of first-order   partial derivatives decreases the accuracy of the estimation process. Authors in \cite{fan2021three} proposed a fast and accurate method for surface normal estimation. The method is also implemented using GPU to achieve higher performance. Despite the superiority of the method, the method is unable to estimate normal vectors for uniform areas.  In our previous work \cite{moradi2021multiple}, a fast and accurate surface normal estimation method is proposed. In that work, a closed-form expression is proposed for each component of the surface normal vectors. Also, the method is capable of multi-scale  implementation which in turn decreases the effect of the measurement noise. However, using multi-scale approach increases the execution time of the algorithm. To address this issue, a fast and accurate surface normal estimation method is presented in this paper. The  contributions of this work are as follows:
 \begin{enumerate}
 \item The averaging process in multi-scale approach is used in multi-direction manner to suppress the effect of measurement noise.
 \item The multi-direction method is implemented in an efficient manner.
 \item The erroneous estimated normal vectors are excluded from final normal map using a simple yet effective method.
 \end{enumerate}
 
The rest of this paper is organized as follows:  In section 2, our previous work is reviewed  briefly as motivation to current work. In section 3 the proposed method is explained in details. Section 4 is dedicated to experiments and results. Finally, the paper is concluded in section 5.

\section{Motivation and Background}

In our previous work \cite{moradi2021multiple}, a fast method was proposed to estimate surface normal vectors directly from depth maps. In this work, first, the projection of two surface tangent vectors in the depth map is constructed (\autoref{fig:surface_tangent_vectors}). Then, the cross product of these two tangent vectors is considered as the surface normal vector at the query point. The closed-form solution for the normal vectors were derived as follows:

\begin{figure}[b!]
     \centering
         \includegraphics[width=0.25\textwidth]{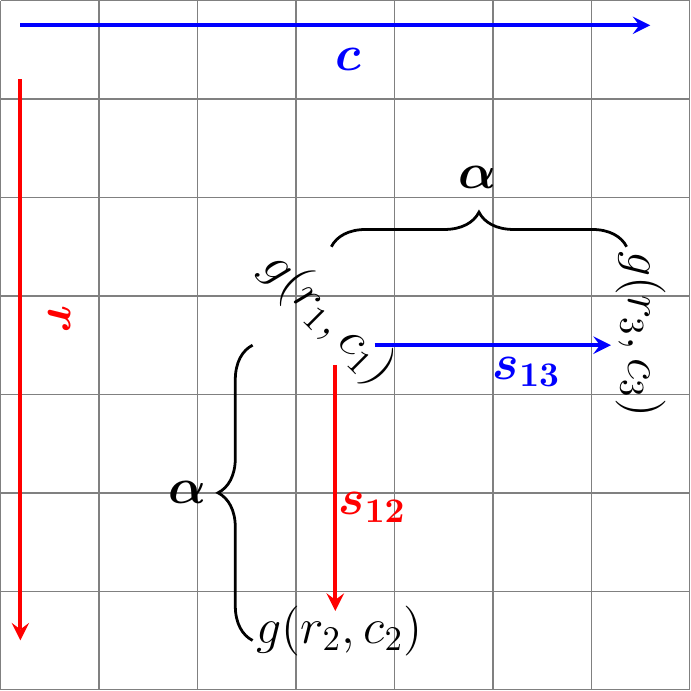}
        \caption{2D projection of surface tangent vectors on the depth map \cite{moradi2021multiple}.}
                \label{fig:surface_tangent_vectors}
\end{figure}
 \begin{equation}
n_x=-\frac{\alpha}{f_y}d_3 \left( d_2 -d_1\right)
\label{eq:nx}
\end{equation}

\begin{equation}
n_y=-\frac{\alpha}{f_x} d_2 \left( d_3 -d_1\right)
\label{eq:ny}
\end{equation}

\begin{equation}
\begin{split}
n_z=&\frac{\alpha}{f_x} v_1d_2 \left( d_3 -d_1   \right)+\frac{\alpha}{f_y} u_1d_3 \left( d_2 -d_1   \right)\\
&+\frac{\alpha^2}{f_xf_y}d_2d_3
\end{split}
\label{eq:nz}
\end{equation}
where, $f_x$ and $f_y$ denote the focal lengths. Also,  $d_i=g(r_i,c_i)$, $u_i=\frac{r_i-o_x}{f_x}$, $v_i=\frac{c_i-o_y}{f_y}$. $o_x$ and $o_y$ are the coordinates of the optical center.

In case of noisy input, the averaging process on multi-scale results will reduce the effect of noise. Therefore:
\begin{equation}
n_x=\frac{-1}{K}\sum_{i=1}^{K}\frac{\alpha_i}{f_y}d_3 \left( d_2 -d_1\right) \quad i=1,2,\cdots,K
\end{equation}
\begin{equation}
n_y=\frac{-1}{K}\sum_{i=1}^{K}\frac{\alpha_i}{f_x} d_2 \left( d_3 -d_1\right) \quad i=1,2,\cdots,K
\end{equation}
\begin{equation}
\begin{split}
n_z=&\frac{1}{K}\sum_{i=1}^{K}\frac{\alpha_i}{f_x} v_1d_2 \left( d_3 -d_1   \right)+\frac{\alpha_i}{f_y} u_1d_3 \left( d_2 -d_1   \right)\\ &+\frac{\alpha_i^2}{f_xf_y}d_2d_3 \quad i=1,2,\cdots,K
\end{split}
\end{equation}

While the single-scale version of this method is fast and gives an accurate estimation of normal vectors for smooth surfaces, the multi-scale normal estimation is slower  by a factor equal to the number of  scales. Moreover, the effect of depth discontinuity at object boundaries is not considered. In the next section, both shortcomings of our previous work are addressed and a fast, accurate, and object boundary-aware surface normal vector estimation method is presented.
\section{The Proposed Method}
\subsection{Fast normal estimation}
\begin{figure}[h!]
     \centering
         \includegraphics[width=0.25\textwidth]{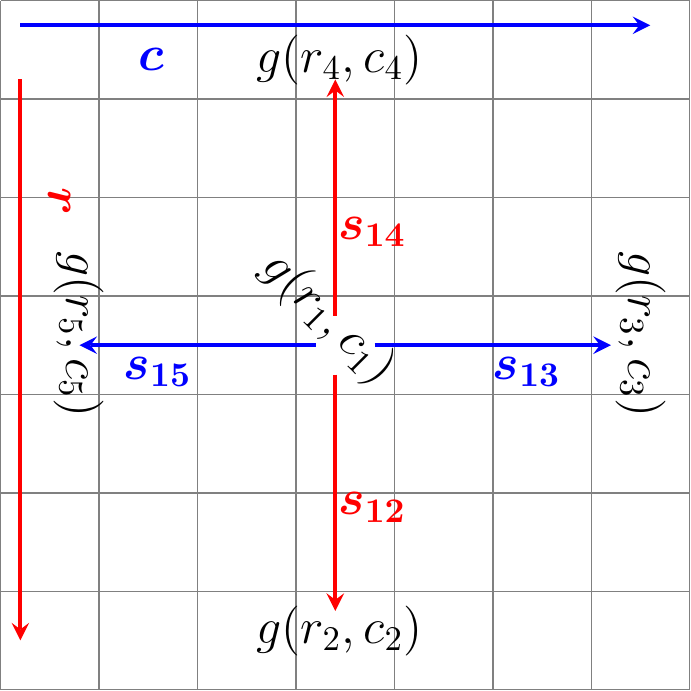}
        \caption{Projection of surface tangent vectors construction in all four main directions}
                \label{fig:surface_tangent_vectors_memar}
\end{figure}

In our previous work \cite{moradi2021multiple}, a multi-scale approach is used to reduce the effect of measurement noise on the final estimated surface normal vectors. The averaging operation among different scales can effectively reduce the noise effect. However, using $K$ scales for final normal construction increases the execution time by a factor of $K$ compared to the single-scale approach. In order to benefit from averaging operation in noise effect reduction, here, we use  multiple pairs of different tangent vectors to estimate the surface normal at a query point. Then, the final normal vector at each point is obtained by taking the average value of the resulting normal vectors.    \autoref{fig:surface_tangent_vectors_memar} shows the projection of four  different surface tangent vectors on depth maps. While, the cross product of  every vectors pair can  be used to  estimate  the normal vectors, only perpendicular tangent vectors pairs are considered, here. The average value of all resulting normal vectors is the final normal vector for each query point. The averaging process makes this estimation robust against measurement noise. 

\begin{figure}[h!]
     \centering
         \includegraphics[width=0.25\textwidth]{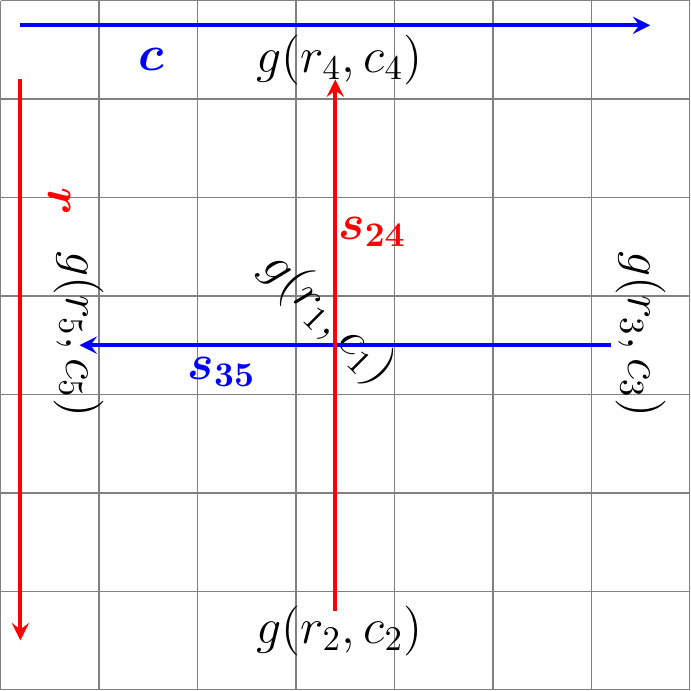}
        \caption{Projection of final surface tangent vectors}
                \label{fig:surface_tangent_vectors_final}
\end{figure}

Considering all four tangent vector pairs, the normal vector can be estimated as: 
\begin{equation}
\begin{split}
n&=0.25(n_{23}+n_{34}+n_{45}+n_{52}) \\
&=0.25(s_{12}\times s_{13}+s_{13}\times s_{14}+s_{14}\times s_{15}+s_{15}\times s_{12}) \\
&=0.25(s_{12}\times s_{13}-s_{14}\times s_{13}+s_{14}\times s_{15}-s_{12}\times s_{15}) \\
&=0.25(s_{12}-s_{14})\times s_{13}+(s_{14}-s_{12})\times s_{15}) \\
&=0.25(s_{12}-s_{14})\times s_{13}-(s_{12}-s_{14})\times s_{15}) \\
&=0.25((s_{12}-s_{14})\times (s_{13}-s_{15}))\\
&=0.25(s_{24}\times s_{35})
\end{split}
\label{eq:proof_of_efficiency}
\end{equation}
where, $\bm{s_{24}}$ and $ \bm{s_{35}}$ are two tangent vectors which are depicted in \autoref{fig:surface_tangent_vectors_final}. \autoref{eq:proof_of_efficiency} proves that the result of the summation of four different cross products can be achieved  using a single cross product. This means that using this approach can accelerate the normal estimation process by a factor of 4.   Finally, the closed-form solution for the normal vectors can be determined as:
\begin{equation}
\bm{s_{24}}=
\begin{bmatrix}
x_4-x_2\\
y_4-y_2\\
z_4-z_2
\end{bmatrix}
=
\begin{bmatrix}
u_4d_4-u_2d_2\\
v_4d_4-v_2d_2\\
d_4-d_2
\end{bmatrix}
\end{equation}
\begin{equation}
\bm{s_{35}}=
\begin{bmatrix}
x_5-x_3\\
y_5-y_3\\
z_5-z_3
\end{bmatrix}
=
\begin{bmatrix}
u_5d_5-u_3d_3\\
v_5d_5-v_3d_3\\
d_5-d_3
\end{bmatrix}
\end{equation}

 \begin{equation}
n_x=-\frac{\alpha}{4f_y}\left( d_3 + d_5\right) \left( d_2 -d_4\right)
\label{eq:nx2}
\end{equation}

\begin{equation}
n_y=-\frac{\alpha}{4f_x} \left( d_2 + d_4\right) \left( d_3 -d_5\right)
\label{eq:ny2}
\end{equation}

\begin{equation}
n_z=-u_1n_x-v_1n_y+ \frac{\alpha^2}{4f_xf_y}\left( d_2 + d_4\right) \left( d_3 +d_5\right)
\label{eq:nz2}
\end{equation}

\subsection{Considering object boundaries}\label{sec:obc}
The proposed method as well as our previous work \cite{moradi2021multiple} works well when facing the points belonging to surfaces without depth discontinuances. However, at the object boundaries, at least one of the  surface tangent vectors is not valid. Therefore, the orientation of the estimated normal vector may be erroneous. To tackle this problem,  a new approach is presented here. 

Surface normals are unit vectors and their orientations are the only important parameters in 3D processing. This is why only $\phi$ and $\theta$ components are taken into account when a normal vector is converted from Cartesian coordinates into spherical one. However, the length of estimated normal vector depends on the length of tangent vectors as fallows:
\begin{equation}
\parallel n\parallel_2 ~\propto~ \parallel s_{24}\times s_{35}\parallel_2 ~\propto~  \parallel  s_{24}\parallel_2 .\parallel s_{35}\parallel_2
\label{eq:veclength}
\end{equation}  

Since at least one of the tangent vectors has a large length in object boundaries,  \autoref{eq:veclength} indicates that the length of the normal also should be large in those areas. Thus, the length of the vector ($r$ component in spherical coordinates) can be used as a mask to find and exclude erroneous normal estimation. A simple thresholding on $r$ values gives the location of outliers. The final normal estimation is performed by applying the invalid points mask to the estimated normal map.
\section{Experimental results}
In order to evaluate the normal estimation performance of the proposed method,  experiments were carried out on real data captured by a Microsoft Kinect Azure
RGB-D camera as well as synthetic depth from 3F2N \cite{fan2021three} dataset.  All the algorithms are implemented in MATLAB. The full specifications of implementation environment are reported in \autoref{tab:spec_runtime}.

There are two ways to demonstrate a surface normal vector as an image. In the first one, each component of the normal vector is considered as a color channel and the resulting vector from depth map can be treated as a color image. While this way is straightforward, it can not highlight small errors in the estimated normal vectors. The second way is to convert the normal vector to spherical coordinates and investigate $I_{\phi}$ and $I_{\theta}$ components. $I_\phi$ and $I_\theta$ can be calculated as:
\begin{equation}
I_\phi=\tan^{-1}\left(\frac{n_y}{n_x}\right)
\end{equation}

\begin{equation}
I_\theta=\tan^{-1} \left( \frac{\sqrt{n_x^2+n_y^2}}{n_z} \right)
\end{equation}

\begin{figure*}[ht!]
\centering
\includegraphics[height=2.95in]{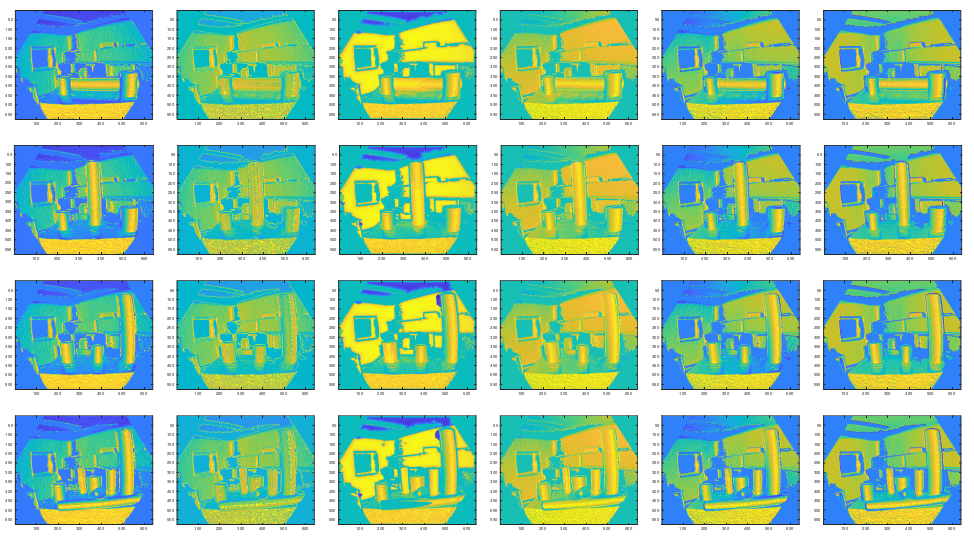}
   	\caption{ $\bm{I_\theta}$ images of the estimation results of different algorithms (Real images). From left: the  ground truth, Fan's method \cite{fan2021three}, Holzer's method \cite{holzer2012adaptive}, Nakagawa's method \cite{nakagawa2015estimating}, Moradi's method \cite{moradi2021multiple}, \textbf{the proposed method}. }
   	\label{fig:normal-estimation_theta}
\end{figure*}
\begin{figure*}[ht!]
\centering
\includegraphics[height=2.9in]{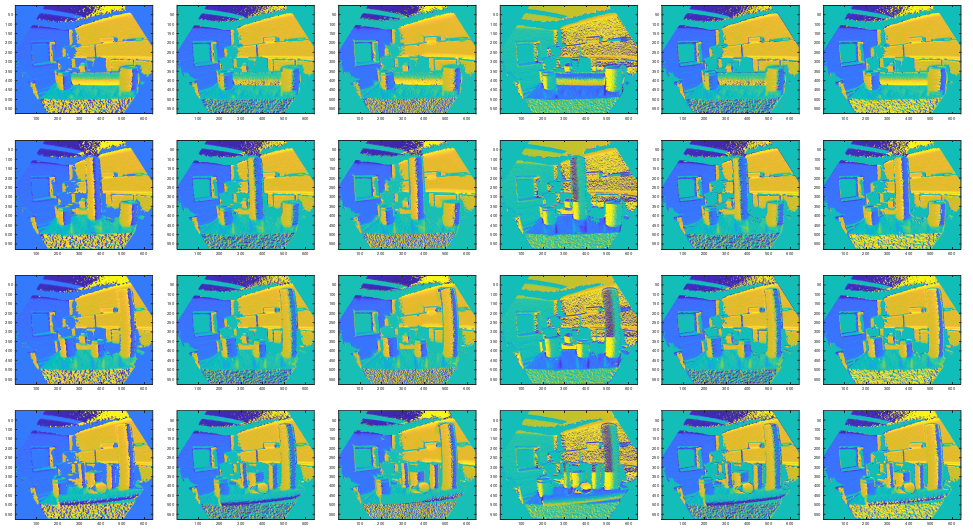}
   	\caption{ $\bm{I_\phi}$ images of the estimation results of different algorithms (Real images). From left: the  ground truth, Fan's method \cite{fan2021three}, Holzer's method \cite{holzer2012adaptive}, Nakagawa's method \cite{nakagawa2015estimating}, Moradi's method \cite{moradi2021multiple}, \textbf{the proposed method}. }
   	\label{fig:normal-estimation_phi}
\end{figure*}
\begin{figure*}[ht!]
\centering
\includegraphics[height=2.8in]{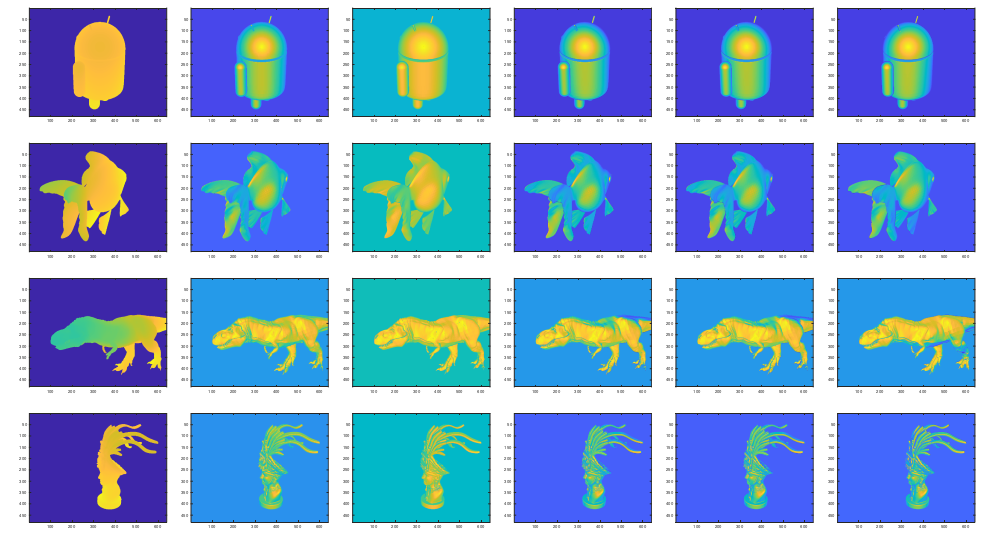}
   	\caption{ $\bm{I_\theta}$ images of the estimation results of different algorithms (Synthetic images from 3F2N dataset \cite{fan2021three}). From left: the depth image, ground truth, Fan's method \cite{fan2021three},  Nakagawa's method \cite{nakagawa2015estimating}, Moradi's method \cite{moradi2021multiple}, \textbf{the proposed method}. }
   	\label{fig:normal-estimation_theta_fan}
\end{figure*}
\begin{figure*}[ht!]
\centering
\includegraphics[height=2.9in]{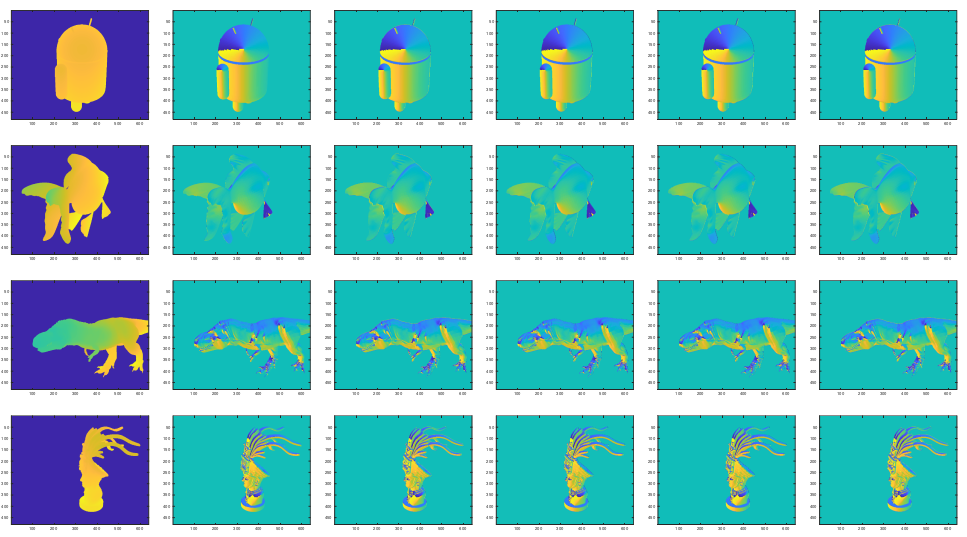}
   	\caption{ $\bm{I_\phi}$ images of the estimation results of different algorithms (Synthetic images from 3F2N dataset \cite{fan2021three}). From left: the depth image, ground truth, Fan's method \cite{fan2021three},  Nakagawa's method \cite{nakagawa2015estimating}, Moradi's method \cite{moradi2021multiple}, \textbf{the proposed method}. }
   	\label{fig:normal-estimation_phi_fan}
\end{figure*}

The results of $I_{\theta}$ and $I_{\phi}$ for different real depth maps are depicted in \autoref{fig:normal-estimation_theta} and \autoref{fig:normal-estimation_phi}, respectively. As shown in these figures, the proposed method outperforms all the baseline algorithms in term of similarity to the ground truth images. The second place belongs to our previous work \cite{moradi2021multiple}. In order to have a fair comparison, the $\alpha$ value for both algorithms is set to $2$.  Also, the results of the plane fitting-based method with $25$ neighboring points are considered as ground truths. \autoref{fig:normal-estimation_theta_fan} and \autoref{fig:normal-estimation_phi_fan} show the result of normal estimation on 3F2N dataset. Again, the proposed method and our previous work outperform the other works.  Since, the synthetic images are smooth enough, the Nakagawa's method does  not affected by sensitivity of partial derivatives to noise. This is why this method performs well on synthetic data compared to real data.

  \begin{table}[!t]
	\centering
	\caption{ The full specifications of the implementation environment}
	\begin{tabular}{|c|c|}
		\hline
	Operating System	 & Ubuntu 20.04 \\
		\hline

		MATLAB version  &  2021b \\

		\hline
				Size of the test image &   576$\times$640 \\
		\hline
				data  type & double precision 64 bit floating point \\
				\hline
		CPU &  Intel CORE i7-3520M  @ 2.90GHz \\
		\hline
		Memory &   16GB DDR3 @ 1600MHz \\
		\hline
		\end{tabular}
	\label{tab:spec_runtime}
\end{table}

For quantitative comparison, the Mean Squared error (MSE) is used as the performance metric.
 \autoref{tab:tabp1}, \autoref{tab:tabp2}, \autoref{tab:tabp1fan}, and \autoref{tab:tabp2fan} show the MSE values of both $\theta$ and $\phi$ image components. As reported in the tables, the proposed method shows the best performance among all the baseline algorithms except for the $\phi$  component of the synthetic depth images which there is not a significant difference among all the baseline algorithms. Since the synthetic data are relatively smooth, the $\alpha$ value for the proposed method as well as our previous work \cite{moradi2021multiple} is set to $1$. In this situation, Nakagawa's method shows same behavior as our previous work (\autoref{tab:tabp1fan} and and \autoref{tab:tabp2fan}).  Note that all images are normalized in $[0-2\pi ]$ range.

\begin{table}[t!]
\centering
\caption{Mean Squared error (MSE) of  $I_\theta$ images}
\label{tab:tabp1}
\begin{tabular}{|c|c|c|c|c|c|} 
\cline{2-6}
\multicolumn{1}{c|}{} & \cite{holzer2012adaptive} & \cite{nakagawa2015estimating} & \cite{fan2021three} & \cite{moradi2021multiple} &ours             \\ 
\hline
$1^{st}$ scene             &  3.2845          & 2.4227 & 1.8098  &    1.0161     & \textbf{0.9301}  \\ 
\hline
$2^{nd}$ scene             &  2.9482          & 2.3099   & 1.3769  &  1.0783     & \textbf{0.9150}  \\ 
\hline
$3^{rd}$ scene             &  2.6559          &  2.6072  & 1.2687  &    1.0244     & \textbf{0.8032}  \\ 
\hline
$4^{th}$ scene             &  2.8190          & 2.7241   & 1.2618  &    1.0012    & \textbf{0.8918}  \\
\hline
\end{tabular}
\end{table}

\begin{table}[t!]
\centering
\caption{Mean Squared error (MSE) of  $I_\phi$ images}
\label{tab:tabp2}
\begin{tabular}{|c|c|c|c|c|c|} 
\cline{2-6}
\multicolumn{1}{c|}{} &  \cite{holzer2012adaptive} & \cite{nakagawa2015estimating}& \cite{fan2021three} & \cite{moradi2021multiple} &ours             \\ 
\hline
$1^{st}$ scene             &  1.8521          & 5.2841  & 1.9734  &     1.8842    & \textbf{1.6655}  \\ 
\hline
$2^{nd}$ scene             &  2.0199          & 5.1293  & 2.0401   &    2.0785   & \textbf{1.7884}  \\ 
\hline
$3^{rd}$ scene             &  2.0313          & 5.1036  & 2.1251  &   2.0288      & \textbf{1.7738}  \\ 
\hline
$4^{th}$ scene             &  2.2078         & 4.8315  & 2.2550   &   2.1463    & \textbf{1.9697}  \\
\hline
\end{tabular}
\end{table}

\begin{table}[t!]
\centering
\caption{Mean Squared error (MSE) of  $I_\theta$ images (3F2N dataset \cite{fan2021three})}
\label{tab:tabp1fan}
\begin{tabular}{|c|c|c|c|c|} 
\cline{2-5}
\multicolumn{1}{c|}{} & \cite{nakagawa2015estimating} & \cite{fan2021three} & \cite{moradi2021multiple} &ours             \\ 
\hline
$1^{st}$ scene             &  0.0790 & 3.3966  &     0.0779     & \textbf{ 0.0491}  \\ 
\hline
$2^{nd}$ scene             &   0.2408   & 2.8881  &  0.2378     & \textbf{0.1333}  \\ 
\hline
$3^{rd}$ scene             &    \textbf{0.1414}  &   0.7437  &   0.1415     & 0.1465  \\ 
\hline
$4^{th}$ scene             &   0.8808   & 0.7672  &   0.8852    & \textbf{0.6874}  \\
\hline
\end{tabular}
\end{table}

\begin{table}[t!]
\centering
\caption{Mean Squared error (MSE) of  $I_\phi$ images (3F2N dataset \cite{fan2021three})}
\label{tab:tabp2fan}
\begin{tabular}{|c|c|c|c|c|} 
\cline{2-5}
\multicolumn{1}{c|}{} & \cite{nakagawa2015estimating}& \cite{fan2021three} & \cite{moradi2021multiple} &ours             \\ 
\hline
$1^{st}$ scene             &  \textbf{0.0363}  & 0.0616  &     \textbf{0.0363}    & 0.0616 \\ 
\hline
$2^{nd}$ scene             &  \textbf{0.0422}  & 0.0517   &    \textbf{0.0422}   & 0.0517  \\ 
\hline
$3^{rd}$ scene             &  \textbf{0.1259} &  0.1411  &   \textbf{0.1259}     &  0.1411 \\ 
\hline
$4^{th}$ scene             &    \textbf{0.1935}  & 0.2142  &    0.1937    & 0.2143 \\
\hline
\end{tabular}
\end{table}

\begin{figure}[ht!]
   	\centering
   	\subfloat[]{\includegraphics[height=1.25in]{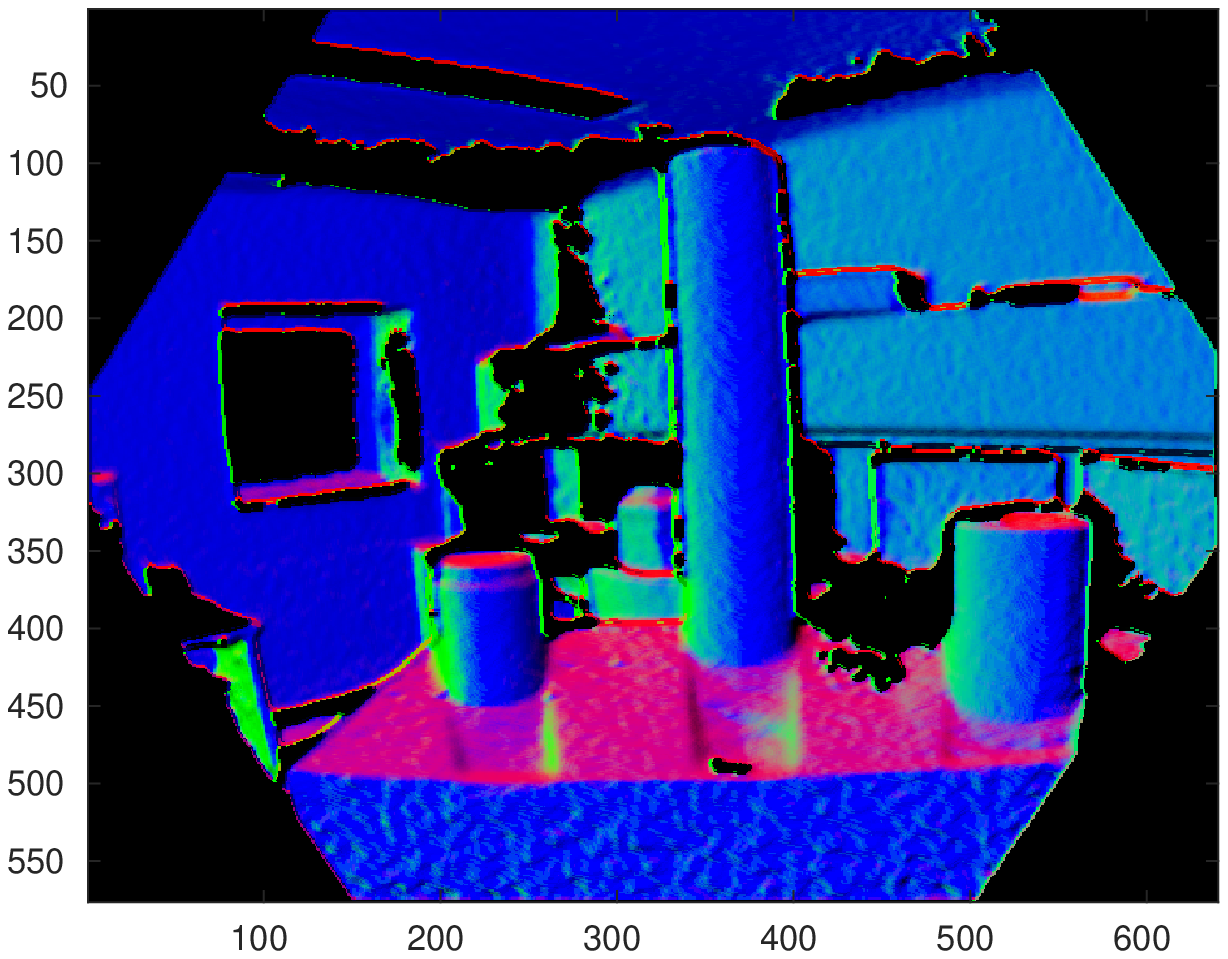}%
   		\label{fig:initnormals}}
   		~
   	   		   	\subfloat[]{\includegraphics[height=1.25in]{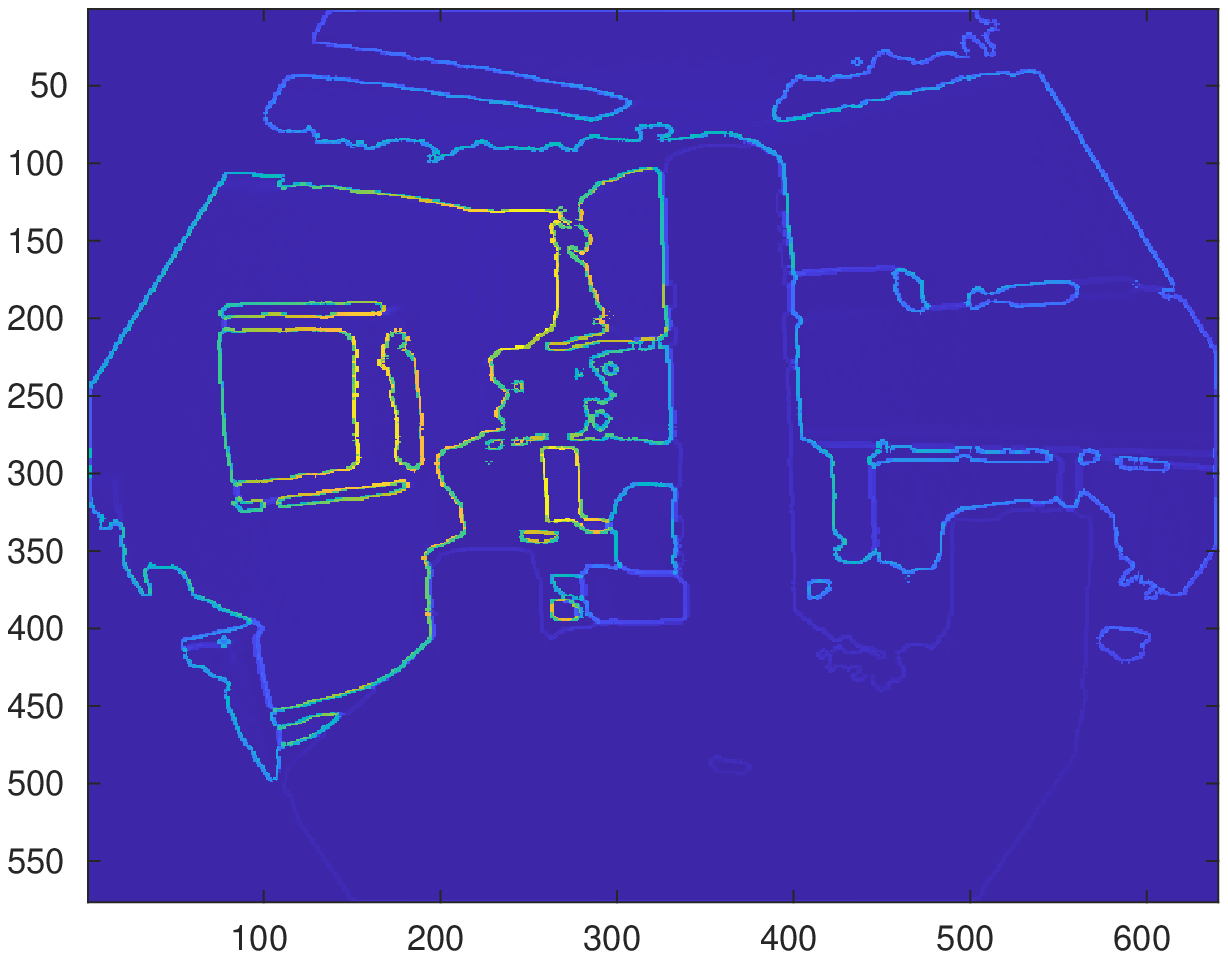}%
   		\label{fig:rcomponent}}
   	\\
   	   	\subfloat[]{\includegraphics[height=1.25in]{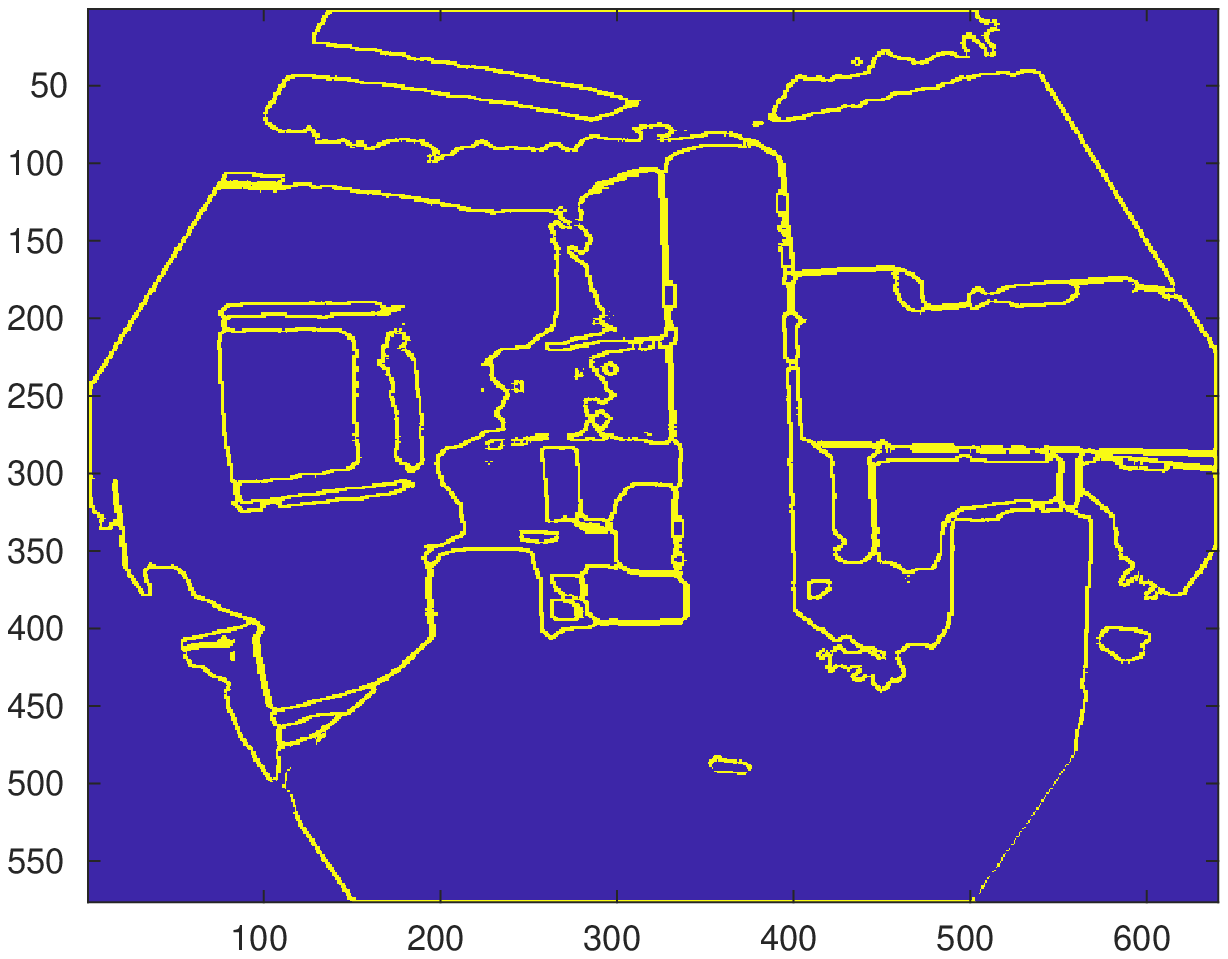}%
   		\label{fig:rmask}}
   		~
   		   	   	\subfloat[]{\includegraphics[height=1.25in]{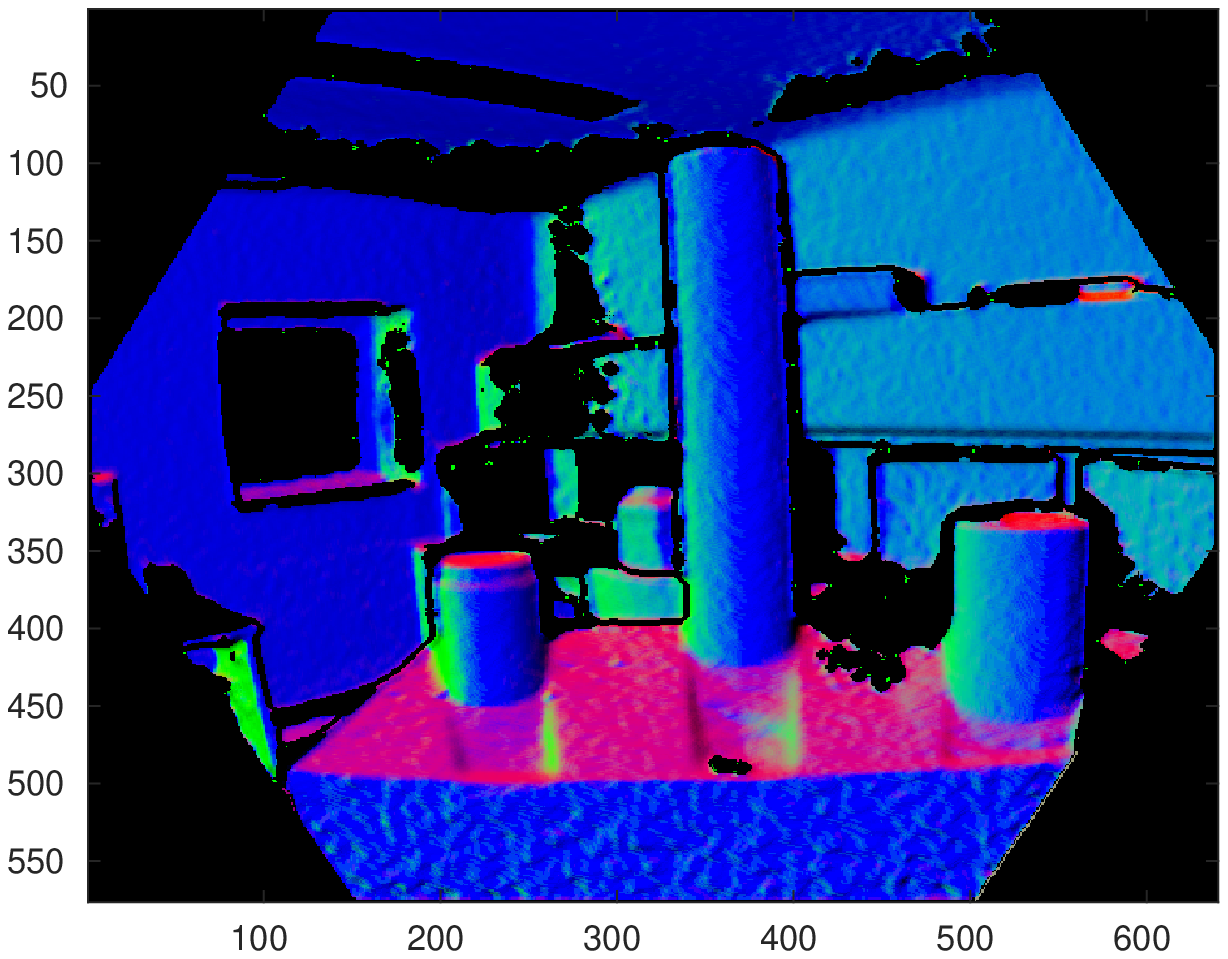}%
   		\label{fig:refindnormalssub}}

   	\caption[]{ Surface normal vectors refinement at object boundaries. a) initial surface normal estimation using \autoref{eq:proof_of_efficiency}, b) value of $r$ component for each point, c) outlier removal mask, and d) final result of surface normal estimation}
   	\label{fig:refinednormals}
   \end{figure}

The color image representation of normal vectors  can be used to demonstrate the erroneous surface normal estimation at object boundaries. \autoref{fig:refinednormals} shows the invalid normal vectors removal operation, step by step. Fig. \autoref{fig:initnormals} shows the estimated surface normal vectors using \autoref{eq:proof_of_efficiency}. As shown in this figure, there are some erroneous estimation at the locations with the depth discontinuity. As mentioned in \autoref{sec:obc}, the large values of $r$ component (Fig. \autoref{fig:rcomponent}) in spherical coordinates have a high potential to be the erroneous estimation. A simple thresholding on $r$ values gives the location of outliers (Fig. \autoref{fig:rmask}). The final normal estimation is performed by applying the mask in Fig. \autoref{fig:rmask} to Fig. \autoref{fig:initnormals}. Fig. \autoref{fig:refindnormalssub} shows the final result of surface normal estimation.

Finally, in order to evaluate the running time of the baseline algorithms as well as the proposed one, all the algorithms are implemented in the MATLAB environment (The MATLAB implementation of the 3F2N method is available in their github repository \cite{fan2021three}). Each algorithm is executed $200$ times and the average running times are reported in \autoref{tab:ms_exe_time}. As reported in the table, the proposed method ranked in second place after Moradi's method (our previous work) \cite{moradi2021multiple}. The proposed method is slightly slower than our previous work. Note that, a single scale implementation of the proposed method is equivalent to four scales implementation of our previous work. Thus, this work is almost four times faster than our previous work.  The normal estimation process can be performed in $135$ fps using the proposed method. Fan's method has a straightforward implementation. However, searching for invalid points ($\Delta Z=0$) in the normal map increases the execution time of this method.
   
\begin{table}[!t]
	\centering
	\caption{The average execution time for different normal estimation methods for a $576\times 640$ depth image}
	\begin{tabular}{|c|c|}
		\hline
	\textbf{Estimation method}	 & \textbf{ execution time (mS)} \\
		\hline
	local plane fitting	 &  $3776$  \\
		\hline
		Holzer's method \cite{holzer2012adaptive} &  $804.620$ \\
				\hline
		Nakagawa's method \cite{nakagawa2015estimating} &   $17.132$ \\
		\hline
		Fan's method \cite{fan2021three} &   $154.676$ \\
		\hline
				Moradi's method \cite{moradi2021multiple} &   $6.646$ \\
		\hline
		\textbf{The proposed method} & \textbf{$7.415$} \\
		\hline
		\end{tabular}
	\label{tab:ms_exe_time}
\end{table}
\section{Conclusion}
In this paper an improved version of our previous work is presented. In our previous work, the normal estimation process is formulated as a closed-form solution which was efficiently implemented. However, multi-scale implementation  is necessary for tackling measurement noise which in turn leads to lower computational efficiency. In this work, instead of changing the pixel distance value parameter $\alpha$ the averaging process is adopted for different directions  to reduce the effects of the measurement noise. Next, the multi-direction approach is reformulated so that it can  be efficiently implemented. So far, the proposed normal estimation method is fast and accurate for smooth surfaces. However, it produces erroneous results when facing object boundaries and depth discontinuities. To address this issues, a simple yest effective mask is constructed based on the length of the estimated normal vectors. This mask is used to exclude erroneous normal vectors from final normal map. 
The proposed method is compared to some well-known surface normal estimation algorithms. Qualitative and quantitative comparisons on real as well as synthetic depth images show that the proposed surface normal estimation method outperforms the baseline algorithms in both terms of accuracy and computational complexity.

{\small
\bibliographystyle{ieee_fullname}
\bibliography{References}
}

\end{document}